\documentclass[10pt,twocolumn,letterpaper]{article}

\usepackage{cvpr}
\usepackage{times}
\usepackage{epsfig}
\usepackage{graphicx}
\usepackage{amsmath}
\usepackage{amssymb}

\usepackage{graphicx}
\usepackage{amsmath,amssymb} 
\usepackage{color}
\usepackage{epsfig}
\usepackage{cite}
\usepackage{bm}
\usepackage{makecell}
\usepackage{booktabs}
\usepackage{multirow}

\usepackage{xspace}
\makeatletter
\DeclareRobustCommand\onedot{\futurelet\@let@token\@onedot}
\def\@onedot{\ifx\@let@token.\else.\null\fi\xspace}
\def\eg{\emph{e.g}\onedot} 
\def\ie{\emph{i.e}\onedot} 
 
\def\etc{\emph{etc}\onedot} 
\def\etal{\emph{et al}\onedot}
\makeatother

\usepackage[pagebackref=true,breaklinks=true,letterpaper=true,colorlinks,bookmarks=false]{hyperref}

\cvprfinalcopy 

\def\cvprPaperID{11816} 

\ifcvprfinal\fi
\begin{document}

\title{High-Resolution Photorealistic Image Translation in Real-Time:\\ A Laplacian Pyramid Translation Network}

\author{\textbf{Jie Liang}$^{1, 2}$\footnotemark[1],\;  \textbf{Hui Zeng}$^{1, 2}$\footnotemark[1]\; and \textbf{Lei Zhang}$^{1,2}$\footnotemark[2]\\
$^1$The HongKong Polytechnic University,\;  $^2$DAMO Academy, Alibaba Group\\
\textit{\{csjliang,\, cshzeng,\, cslzhang\}@comp.polyu.edu.hk}\\
}

\maketitle

\renewcommand{\thefootnote}{\fnsymbol{footnote}}
\footnotetext[1]{Equal contribution.}
\footnotetext[2]{Corresponding author. This work is supported by the Hong Kong RGC RIF grant (R5001-18).}

\begin{abstract}
Existing image-to-image translation (I2IT) methods are either constrained to low-resolution images or long inference time due to their heavy computational burden on the convolution of high-resolution feature maps. In this paper, we focus on speeding-up the high-resolution photorealistic I2IT tasks based on closed-form Laplacian pyramid decomposition and reconstruction. Specifically, we reveal that the attribute transformations, such as illumination and color manipulation, relate more to the low-frequency component, while the content details can be adaptively refined on high-frequency components. We consequently propose a Laplacian Pyramid Translation Network (LPTN) to simultaneously perform these two tasks, where we design a lightweight network for translating the low-frequency component with reduced resolution and a progressive masking strategy to efficiently refine the high-frequency ones. Our model avoids most of the heavy computation consumed by processing high-resolution feature maps and faithfully preserves the image details. Extensive experimental results on various tasks demonstrate that the proposed method can translate 4K images in real-time using one normal GPU while achieving comparable transformation performance against existing methods. Datasets and codes are available: \href{https://github.com/csjliang/LPTN}{https://github.com/csjliang/LPTN}.
\end{abstract}

\section{Introduction}
	
	Image-to-image translation (I2IT, ~\cite{zhang2019harmonic, hoffman2017cycada, ma2018exemplar}), which aims to translate images from a source domain to a target one, has gained significant attention. Recently, photorealistic I2IT has been attracting increasing interest in various practical tasks, \eg, transferring images among different daytimes or seasons~\cite{hoffman2017cycada} or retouching the illumination and color of images to improve their aesthetic quality ~\cite{chen2018deep}. Different from the general I2IT problem, the key challenge of the practical photorealistic I2IT task is to keep efficiency and avoid content distortions when handling high-resolution images.

	\begin{figure}[t]
		\centering
		\includegraphics[width=0.46\textwidth]{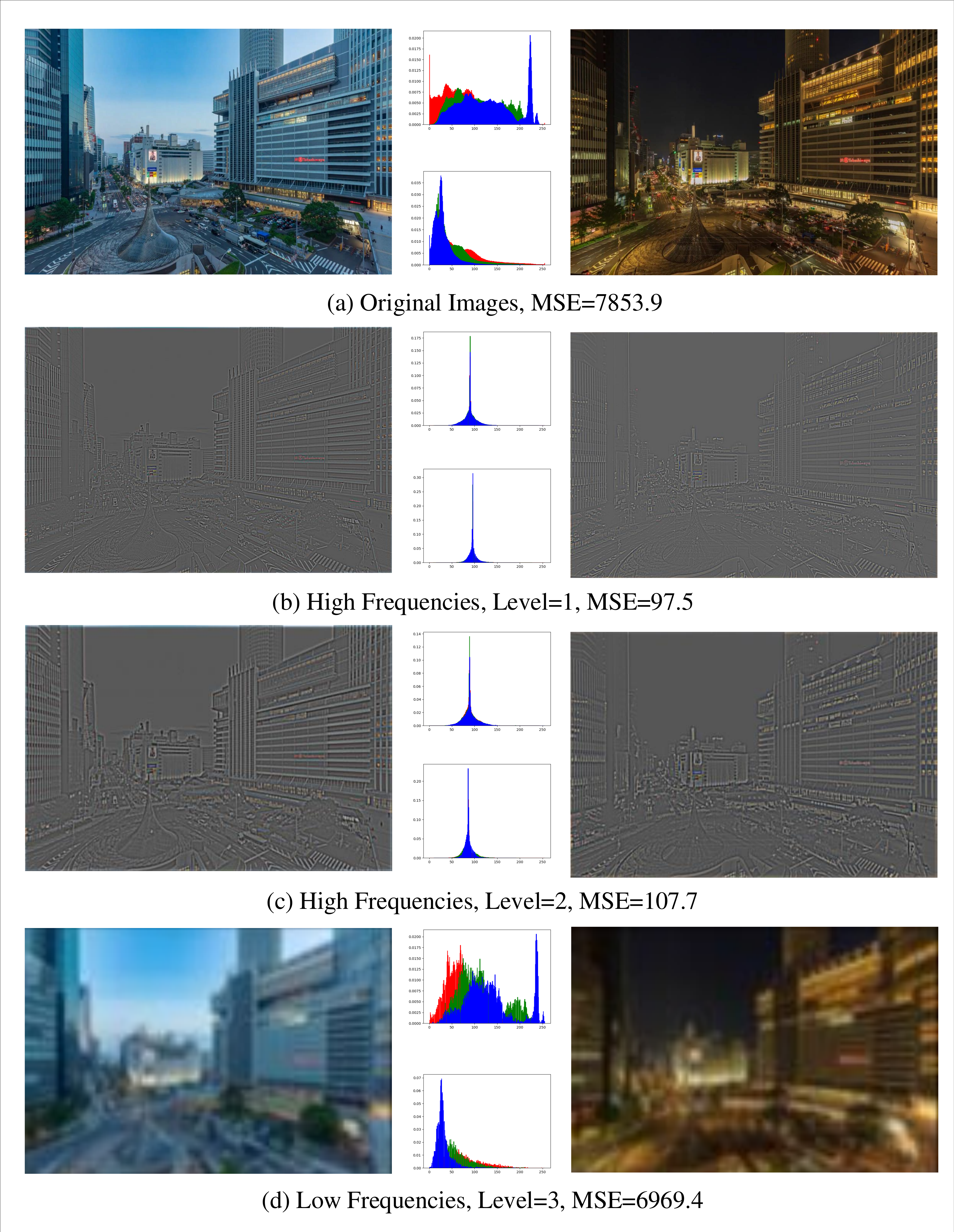}
		\caption{(a) Images of a scene captured at different daytimes and (b$\sim$d) the Laplacian pyramids (figures in (c$\sim$d) are resized for better visualization).
			As shown by the MSE and the Histograms, the differences between the day and night images are dominated in the low-frequency components (d).
			\label{motivation}
		}
	\end{figure}
	
	To achieve faithful translations, most traditional methods~\cite{zhu2017unpaired, wang2018high, isola2017image} employ an encoding-decoding paradigm which maps the input image into a low-dimensional latent space, followed by reconstructing the output from a translated latent code. However, these methods are naturally limited to low-resolution applications or time-consuming inference models~\cite{zhu2017unpaired, wang2018high, isola2017image, li2018closed, luan2017deep, lee2018diverse}, which is far from practical. The main reason is that the model needs to manipulate the image globally using deep networks, yet directly convolving a high-resolution image with sufficient channels and large kernels demands heavy computational cost. There are some developments in pruning and boosting the inference models~\cite{johnson2016perceptual, huang2017real, li2019learning}, yet a shallow network can hardly fulfill the requirements of reconstructing complex content details from a low-dimensional latent space to a high-resolution image. To generate a photorealistic translation, recent researches~\cite{gonzalez2018image, huang2017arbitrary, huang2018multimodal} have also been focusing on disentangling the contents and attributes of both domains in a data-driven manner. Nevertheless, the irreversible down- and up-sampling operations in these models still involves heavy convolutions on high-resolution feature maps, sacrificing the efficiency of the inference model.

	Inspired by the reversible and closed-form frequency-band decomposition framework of a Laplacian pyramid (LP,~\cite{burt1983laplacian}), we reveal that the domain-specific attributes, \eg, illuminations or colors, of a photorealistic I2IT task are mainly exhibited on the low-frequency component. In contrast, the content details relate more to higher-frequency components, which can be adaptively refined according to the transformation of the visual attributes. As shown in Figure~\ref{motivation}, for a pair of images with the same scene yet captured at different daytimes, the mean squared errors (MSE) between the high-frequency components (b-c) of the two domains are much smaller (about $1/71$ and $1/65$) than that between the low-frequency components (d). Similar findings can be observed from the histograms and visual appearance. Figure~\ref{motivation} (b-c) also demonstrate that the higher-frequency subimages are with tapering resolutions, while different levels show pixel-wise correlations and exhibit similar textures. Such properties allow an efficient masking strategy for adjusting the content details accordingly.
	
	Based on the above observations, in this paper, we propose a fast yet effective method termed the \textit{Laplacian Pyramid Translation Network} (LPTN) to improve efficiency while keeping the transformation performance for photorealistic I2IT tasks. In specific, we build a lightweight network with cascaded residual blocks on top of the low-frequency component to translate the domain-specific attributes. To fit the manipulation of the low-frequency component and reconstruct the image from an LP faithfully, we refine the high-frequency components adaptively yet avoid heavy convolutions on high-resolution feature maps to improve the efficiency. Therefore, we build another tiny network to calculate a mask on the smallest high-frequency component of the LP and then progressively upsample it to fit the others. The framework is trained end-to-end in an unsupervised manner via adversarial training strategy.
	
	The proposed method offers multiple advantages. Firstly, we are the first to enable photorealistic I2IT on 4K resolution images in real-time. Secondly, given the lightweight and fast inference model, we still achieve comparable or superior performance on photorealistic I2IT applications in terms of transformation capacity and photorealism. Both qualitative and quantitative results demonstrate that the proposed method performs favorably against state-of-the-art methods.
	
	\begin{figure*}[t]
		\centering
		\includegraphics[width=0.93\textwidth]{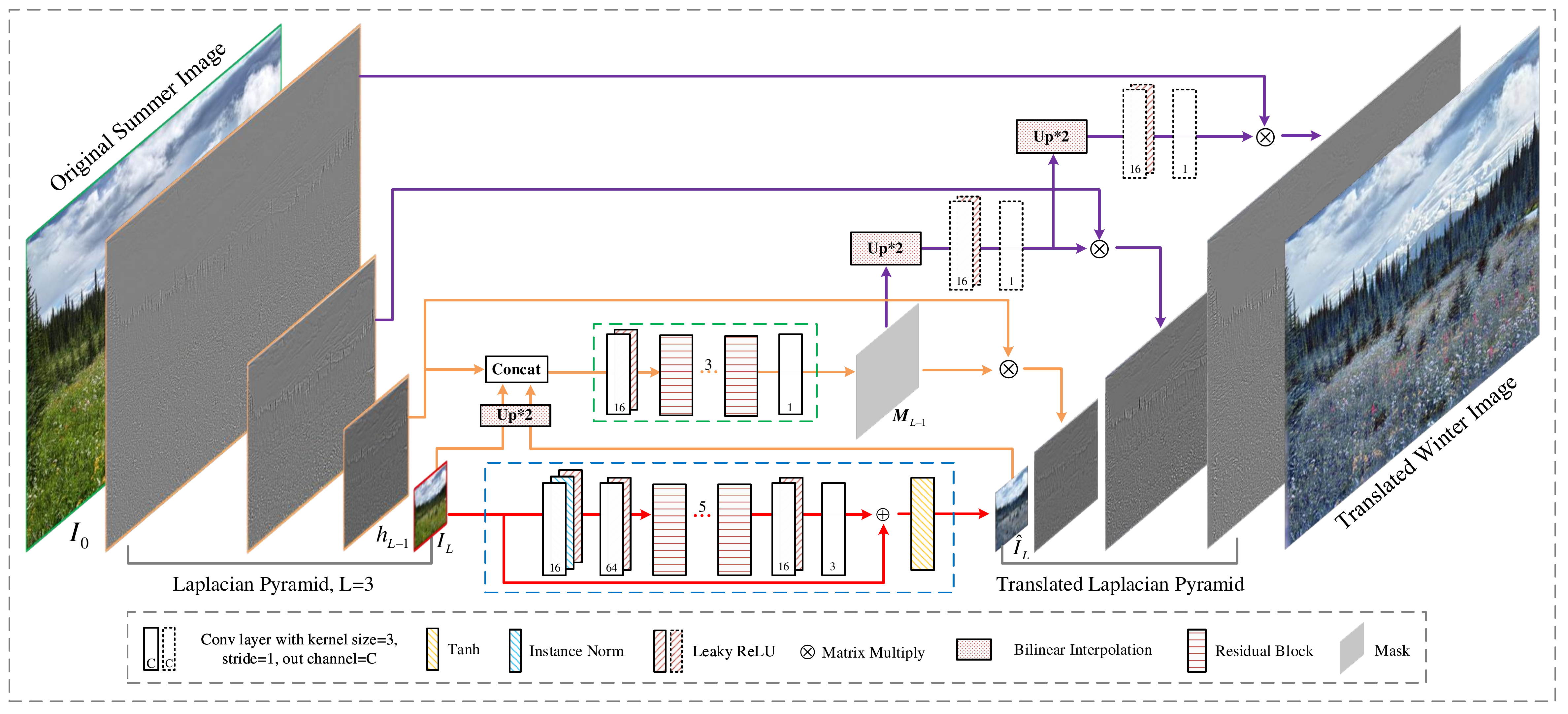}
		\caption{Pipeline of the proposed LPTN algorithm.
			Given a high-resolution image $ I_0\in\mathbb{R}^{h\times w\times 3}$, we first decompose it into a Laplacian pyramid (\eg, $L=3$). \textcolor[rgb]{1,0,0}{Red} arrows: For the low-frequency component $I_L\in\mathbb{R}^{\frac{h}{2^{L}}\times \frac{w}{2^{L}}\times c}$, we translate it into $\hat{I}_L\in\mathbb{R}^{\frac{h}{2^{L}}\times \frac{w}{2^{L}}\times c}$ using a lightweight network. \textcolor[rgb]{0.96,0.64,0.39}{Brown} arrows: To adaptively refine the high-frequency component $h_{L-1} \in \mathbb{R}^{\frac{h}{2^{L-1}}\times \frac{w}{2^{L-1}}\times c}$, we learn a mask $\bm{M}_{L-1}\in\mathbb{R}^{\frac{h}{2^{L-1}}\times \frac{w}{2^{L-1}}\times 1}$ based on both high- and low-frequency components. \textcolor[rgb]{0.5,0,0.5}{Purple} arrows: For the other components with higher resolutions, we progressively upsample the learned mask and finetune it with lightweight convolution blocks to maintain the capacity of a photorealistic reconstruction.
			\label{pipeline}}
	\end{figure*}

	\section{Related Work}
	
	\subsection{Photorealistic Image Translation}
	
	Most existing I2IT methods~\cite{isola2017image, gonzalez2018image, lee2018diverse, zhu2017unpaired, liu2017unsupervised, zhu2017toward} include three main steps as follows: 1) encoding the image into a low-dimensional latent space; 2) translating the domain-specific attributes in the latent space and 3) reconstructing the image via a deep decoder. Recent researchers attempt to alleviate the space burden and improve the time efficiency of the I2IT models~\cite{johnson2016perceptual, gharbi2017deep, zhang2017real, huang2017real, chen2017coherent, li2019learning, wang2018high}. For example, to allow translation on high-resolution images, Wang~\etal~\cite{wang2018high} proposed a coarse-to-fine generation pipeline where a low-resolution translation is learned first and then expanded progressively to higher-resolution. However, it is computationally expensive due to the direct optimization of high-resolution images. There are also some speeding-up frameworks in the photorealistic style transfer community. Specifically, instead of conducting iterative forward and backward processes~\cite{gatys2016image}, researchers proposed to learn a feed-forward network to approximate the optimization process~\cite{johnson2016perceptual,chen2017coherent,huang2017real}. Nevertheless, the encoding and decoding steps may introduce structural distortions due to the trade-off between efficiency and effectiveness.
	
	To enhance the faithfulness of a fast stylization, Li~\etal~\cite{li2019learning} took advantage of the spatial propagation network~\cite{liu2017learning}, which however can hardly be extended to high-resolution applications. Recent developments~\cite{Disentangler_NeurIPS2018, huang2018multimodal, lee2018diverse, gonzalez2018image, ganin2014unsupervised, yoo2019photorealistic, puy2019flexible} also focus on disentangling the factors of data variations based on second-order statistics. For example, Huang~\etal~\cite{huang2017arbitrary} proposed an adaptive instance normalization which normalizes the content latent code using the mean and standard deviation of the style. To allow a photorealistic translation according to a given reference, Luan~\etal~\cite{luan2017deep} designed a novel loss on preserving the local structure of the given content image. In addition, Li~\etal~\cite{li2018closed} proposed a smoothing process based on per-pixel affinities on top of the original transformation stage. Furthermore, Yoo~\etal~\cite{yoo2019photorealistic} introduced a wavelet pooling strategy to approximate the average pooling yet with a mirroring unpooling operation. Nevertheless, these methods are computationally expensive on high-resolution tasks, \eg, costing a few seconds on an HD image. In addition, they need a reference image to manipulate the style of each input. In contrast, the I2IT methods including the proposed LPTN models the visual attributes based on the overall distribution of the training data, which thus need only the input image in the testing stage.

	\subsection{Laplacian Pyramid}
	\label{relatedwork_laplacian}
	
	Laplacian pyramid (LP)~\cite{burt1983laplacian} is a long-standing technique on image processing. The main idea of the LP method~\cite{burt1983laplacian} is to linearly decompose an image into a set of high- and low-frequency bands, from which the original image can be exactly reconstructed. In specific, given an arbitrary image $I_0$ of $h\times w$ pixels, it firstly calculates a low-pass prediction $ I_1\in\mathbb{R}^{\frac{h}{2}\times \frac{w}{2}} $ where each pixel is a weighted average of the neighboring pixels based on a fixed kernel. To allow a reversible reconstruction, the LP records the high-frequency residual $ h_0 $ as $ h_0 = I_0 - \hat{I}_0 $, where $ \hat{I}_0 $ denotes the upsampled image from $I_1$. To further reduce the sample rate and image resolution, LP iteratively conducts the above operations on $I_1$ to get a sequence of low- and high-frequency components.
	
	The hierarchical structure of the LP paradigm inspires several recent CNN-based image processing works such as image generation~\cite{denton2015deep}, super-resolution~\cite{lai2017deep} and semantic segmentation~\cite{ghiasi2016laplacian}. For example, in order to generate high-quality images, Denton~\etal~\cite{denton2015deep} trained multiple generators on the components of an LP. In addition, Lai~\etal~\cite{lai2017deep} follows the Laplacian pyramid reconstruction process to progressively reconstruct the high-frequency (also high-resolution) components for image super-resolution. Its computation and memory cost grows dramatically with the increase of resolution due to the intensive convolutions on high-resolution components. In contrast, we tackle the photorealistic I2IT problem and reveal that the task can be done by simultaneously translating the illuminations and colors at low-freq and refining slightly the details at high-freq to avoid computationally intensive convolutions. Accordingly, an efficient refining module on high-freq components is designed, allowing a real-time implementation on 4K images.

	\section{Laplacian Pyramid Translation Network}
	
	\subsection{Framework Overview}
	\label{laplacian_pyramid}
	
	We propose an end-to-end framework, namely the Laplacian Pyramid Translation Network (LPTN), to reduce the computational burden and simultaneously keep the transformation performance for photorealistic I2IT tasks. The pipeline of the proposed LPTN is shown in Figure~\ref{pipeline}.
	
	As shown in the figure, given an image $I_0\in\mathbb{R}^{h\times w\times 3}$, we first decompose it into an Laplacian pyramid, obtaining a set of band-pass components denoted by $H = [h_0, h_1, \cdots, h_{L-1}]$ and a low-frequency residual image $I_L$, where $L$ is the number of decomposed levels of the LP. The components of $H$ have tapering resolutions from $h\times w$ to $\frac{h}{2^{L-1}}\times \frac{w}{2^{L-1}}$, while $I_L$ has $\frac{h}{2^{L}}\times \frac{w}{2^{L}}$ pixels. Such a decomposition is invertible where the original image can be reconstructed by a sequence of mirror operations. According to Burt and Adelson~\cite{burt1983laplacian}, $H$ is highly decorrelated where the light intensity of most pixels is close to $0$ except for the detailed textures of the image. At the same time, the low-pass filtered $I_L$ is blurred where each pixel is averaged by the neighboring pixels via an octave Gaussian filter. As a result, $I_L$ reflects the global attributes of an image in a content-independent manner.
	
	Inspired by the above properties of LP, we propose to translate mainly on $I_L$ to manipulate the illuminations or colors, while refining $H$ adaptively to avoid artifacts in reconstruction. In addition, we progressively refine the higher-resolution component conditioned on the lower-resolution one. The LPTN framework is therefore composed of three parts. First, we translate the low-resolution $I_L$ into $\hat{I}_L$ using deep convolutions. Second, we learn a mask on top of the concatenation of $[h_{L-1}, up(I_L), up(\hat{I}_L)]$, where $up(\cdot)$ denotes a bilinear upsampling operation. The mask is then multiplied to $ h_{L-1} $ to refine the high-frequency component of level $ L-1 $. Third, to further refine the other components with higher resolutions, we propose an efficient and progressive upsampling strategy. At each level from $ l=L-2 $ to $l=0$, we first upsample the mask of the last level and then learn a lightweight convolution to slightly finetune the mask. We introduce these modules in detail in the following sections.
	
	\subsection{Translation on Low-Frequency Component}
	\label{low_frequency}
	
	The inherent properties of LP, including the separation of textures and visual attributes, and the capability of a reversible reconstruction, can benefit the photorealistic I2IT task. For general I2IT tasks with texture manipulations, the domain-specific attributes are represented in the latent space powered by a deep encoding-decoding network. In contrast, for the task of photorealistic I2IT, we observe that the domain-specific attributes are mainly about illuminations or colors, which can be extracted using fixed kernels in an efficient way. As shown in Figure~\ref{motivation}, for example, the domain-specific visual attributes of the day-to-night translation task are mainly exhibited in the low-frequency component, while the high-frequency ones relate more to the textures. Consequently, we can translate the domain-specific attributes on the low-frequency component with a downscaled resolution, reducing largely the computational complexity against the general I2I methods.
	
	As shown in Figure~\ref{pipeline}, given $I_L$ with a reduced resolution, we first extend the feature map channel-wisely using a $1\times 1$ convolution. Then, we stack $5$ residual blocks on top of the extended feature map. For each residual block, two convolutions with kernel size being $ 3 $ and stride being $ 1 $ are conducted, each is followed by a leaky ReLU. After that, we reduce the channels of the feature maps back to $c$ to get the translated results $\hat{I}_L$, where $c$ denotes the number of channels of the given image. The output is finally added to the original inputs followed by a Tanh activation layer.
	
	Traditional I2IT algorithms also conduct transformation at a low-dimensional space via a cascade of residual blocks. However, the proposed model shows advantages against these methods in the following ways. 1) On time and space efficiency: The decomposition of high- and low-frequency components in an LP is based on a fixed kernel and a simple convolution operation, it is therefore efficient and free of learning from images. Such a strategy is based on a prior knowledge that the photorealistic I2IT task requires to manipulate illuminations and colors while slightly refining the textures accordingly. In contrast, traditional methods access to the low-dimensional latent space via auto-encoders with heavy convolutions on the whole image, which limits their applications to high-resolution tasks. 2) On the disentanglement and reconstruction effectiveness: The separation of different frequency bands in an LP is simple and effective for disentangling and reconstructing an image, as shown in Figure~\ref{motivation}. In contrast, a learning-based auto-encoder in general methods may suffer from a trade-off between the model size and disentanglement/reconstruction effectiveness.
		
	\subsection{Refinement of High-Frequency Components}
	
	To allow a faithful reconstruction when manipulating domain-specific attributes, the high-frequency components $H = [h_0, h_1, \cdots, h_{L-1}]$ should also be refined according to the transformation from $I_L$ to $\hat{I}_L$. In this section, we propose to learn a mask for $ h_{L-1} $ and progressively expand the mask to refine the rest of high-frequency components according to the intrinsic characteristic of LP.
	
	According to the analysis in Section~\ref{laplacian_pyramid}, we have $ h_{L-1} \in \mathbb{R}^{\frac{h}{2^{L-1}}\times \frac{w}{2^{L-1}}\times c} $ and $I_L, \hat{I}_L\in \mathbb{R}^{\frac{h}{2^{L}}\times \frac{w}{2^{L}}\times c}$. We first upsample $I_L$ and $\hat{I}_L$ with bilinear operations to match the resolution of $ h_{L-1} $. Then, we concatenate $[I_L, \hat{I}_L, h_{L-1}]$ and feed it into a tiny network with the same architecture as shown in Figure~\ref{pipeline}. The output channel of the last convolution layers is set to $1$ in this network.
	
	The output of the network $\bm{M}_{L-1}\in\mathbb{R}^{\frac{h}{2^{L-1}}\times \frac{w}{2^{L-1}}\times 1}$ is considered as a per-pixel mask of the $ h_{L-1} $. As shown in Figure~\ref{motivation}, for image pairs in two domains, the high-frequency components on the same level only differ slightly in terms of the global brightness. Therefore, the masks can be interpreted as a global adjustment which is relatively easier to be optimized than the mixed-frequency images. Consequently, we refine the $ h_{L-1} $ by:
	\begin{equation}
	\label{refine}
	\hat{h}_{L-1} =  h_{L-1} \otimes \bm{M}_{L-1},
	\end{equation}
	where $\otimes$ denotes the pixel-wise multiplication.
	
	We then progressively upsample the per-pixel mask $\bm{M}_{L-1}$ to a set of masks $[\bm{M}_{L-2}, \cdots, \bm{M}_1, \bm{M}_0]$ with resolutions from $\frac{h}{2^{L-2}}\times \frac{w}{2^{L-2}}\times 1$ to $h\times w\times 1$ to match the rest high-frequency components. As shown in Figure~\ref{pipeline}, $\bm{M}_{L-1}$ is expanded with a scale factor of $2$ using bilinear interpolation, followed by an optional lightweight convolution block for fine-tuning. The result of this stage, \ie, $\bm{M}_{L-2}$, is then progressively upsampled until $ \bm{M}_0 $ is generated. Consequently, we can refine all the high-frequency components of the LP using the same operations as in Eq.~\eqref{refine} and get the result set $[\hat{h}_0, \hat{h}_1, \cdots, \hat{h}_{L-1}]$. The result image $ \hat{I}_0 $ is then reconstructed using the translated $\hat{I}_L$ and the refined $[\hat{h}_0, \hat{h}_1, \cdots, \hat{h}_{L-1}]$.
	
	To demonstrate the effectiveness of the bilinear interpolation on upsampling the masks, let's recap the construction of an LP. As mentioned in Section~\ref{relatedwork_laplacian}, given the low-frequency image of the $l$-th level, \ie, $I_l$, we have $h_l = I_l - T(C(I_l))$ where $C$ and $T$ denote convolution and transpose convolution with the same low-pass kernel. On the next level, we have $h_{l+1} = I_{l+1} - T(C(I_{l+1})) = C(I_l) - T(C(C(I_0)))$ since $I_{l+1} = C(I_l)$. The closed-form convolution operation $C$ with the 2D low-pass kernel derived from $[1, 4, 6, 4, 1]$ approximates the average pooling with a receptive field of $5$. Figure~\ref{motivation} demonstrates that the difference between the high-frequency components of the two images is small and only the global tone has a big difference. As a result, a bilinear upsampling and a lightweight convolution are capable to simultaneously reverse the down-sampling process and manipulate the global intensity of the mask. Compared with those directly convolute the large-scale high-frequency components, the above mentioned progressive masking strategy can save computational resources to a large extent.
	
	\subsection{Learning criteria}
	
	The proposed LPTN is trained in an unsupervised scenario by optimizing a reconstruction loss $ \mathcal{L}_{recons} $ as well as an adversarial loss $ \mathcal{L}_{adv} $ on the image space. To encourage a faithful translation and refinement, we let $ \mathcal{L}_{recons} =\lVert I_0 - \hat{I}_0 \rVert_2^2 $ given the input image $I_0$ and the translated result $\hat{I}_0$. Besides, the $ \mathcal{L}_{adv} $ is computed based on the LSGAN objective~\cite{mao2017least} and a multi-scale discriminator~\cite{wang2018high} to match the target distribution. Specifically, we train the generator $G$ (including both low- and high- frequency modules) to minimize $E_{I_0\sim p_{data}(I_0)}[D(G(I_0)-1)^2]$, and train a discriminator $D$ to minimize $ E_{\widetilde{I}_0\sim p_{data}(\widetilde{I}_0)}[(D(\widetilde{I}_0)-1)^2] + E_{I_0\sim p_{data}(I_0)}[D(G(I_0))^2] $. Like~\cite{mao2017least}, the $D$ has 3 components with identical network structure on 3 image scales. The total loss is calculated as follows: $\mathcal{L} = \mathcal{L}_{recons}+\lambda\mathcal{L}_{adv}$, where $\lambda$ balances the two losses.

	\section{Experiment}
	
	\subsection{Setup}
	\label{tasks}
	
	\noindent\textbf{Datasets:} To extend the I2IT task to a high-resolution scenario, we collect two unpaired datasets from Flickr\footnote{https://www.flickr.com/} with random resolutions from 1080p ($1920\times 1080$) to 4K ($3840\times 2160$). One of them is regarding the day$ \rightarrow $night translation task (with $ 1035 $ day photos and $ 862 $ night photos) while the other is about the summer$ \rightarrow $winter translation task (with $ 1173 $ summer photos and $ 1020 $ winter photos). Examples of the training images are shown in the supplementary material.
	
		In addition, to quantitatively evaluate the proposed method, we conduct experiments on the MIT-Adobe fiveK dataset~\cite{bychkovsky2011learning} which contains $5,000$ untouched images and the corresponding manually-retouched targets given by photographic experts. We use the targets given by expert C following the existing works~\cite{chen2018deep}, while we employ $4,500$ images for training and the rest $500$ pairs for evaluation. Note we only use the paired samples to calculate the quantitative metrics in the testing stage.

	\begin{figure*}[t]
		\centering
		\includegraphics[width=0.9\textwidth]{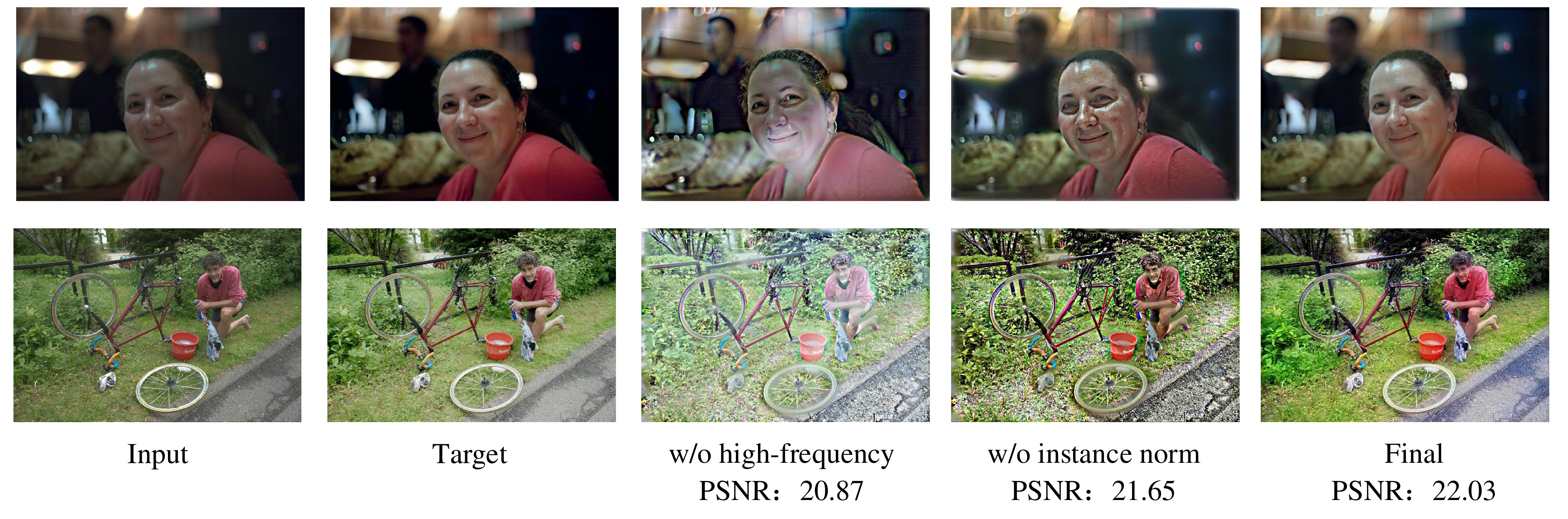}
		\caption{Ablation study toward the model structures on the photo retouching task. The images in the third column are generated without the refinement modules of the high-frequency components, while the images in the fourth column are generated by removing the instance norm layer when translating the low-frequency component. The PSNRs are the average of $500$ test images under the specific setting.
			\label{ablation_model}}
	\end{figure*}

	\noindent\textbf{Hyper-Parameters:} We use an Adam optimizer with the learning rate being $1e^{-4}$. The weight of the losses is set to be $ \mathcal{L}_{recons} : \mathcal{L}_{adv}  = 10:1 $.
	
	\noindent\textbf{Compared Methods:} We compare our method with both unpaired I2IT methods, \ie, CycleGAN~\cite{zhu2017unpaired}, UNIT~\cite{liu2017unsupervised} and MUNIT~\cite{huang2018multimodal} and unpaired photo retouching methods, \ie, the White-box~\cite{hu2018exposure} and DPE~\cite{chen2018deep}. Qualitative and quantitative comparisons are reported in Section~\ref{visual_comp} and Section~\ref{quantitative_comp}, respectively.

	\subsection{Ablation Study}
	
	\noindent\textbf{Effectiveness of Specific Modules: } We visualize the effectiveness of different modules (the refinement of high-frequency components and the instance normalization when translating the low-frequency component) in Figure~\ref{ablation_model}. On one hand, as shown in the third column of the figure, the progressive refinement of the high-frequency components is effective in preserving the texture details. When we remove these refinement modules, although the visual attributes (in this task, illuminations and colors, \etc) are successfully translated, many regions suffer from blurring effects and the PSNR is thus reduced to $20.87$. This is mainly caused by the dis-match between the translated low-frequency component and the nearly unchanged high-frequency ones. On the other hand, as shown in the fourth column of the figure, the instance norm is required when translating the low-frequency component. If we manipulate the attributes with no normalization process, the translation will be excessive and lead to over-sharpened results. As shown in the top row, many undesired details on the face are produced. In contrast, LPTN achieves a natural and photorealistic translation, which results in a comparable PSNR with the state-of-the-art unpaired photo retouching methods.
	
	\noindent\textbf{Selection of the Number of Levels: } We validate the influence of the number of levels $L$ on the photo retouching task. As shown in the last three rows of Table~\ref{psnr_enhance}, the model achieves the best performance on all tested resolutions when $L=3$. At the same time, as shown in the Table~\ref{time_comparison}, the LPTN consumes more time with $L=3$ than that with $L=4$ or $L=5$. Actually, there is a trade-off between the time consumption and the performance, which is determined by the number of levels of the LP. However, the proposed LPTN is robust when increasing the parameter $L$ to reduce the computational burden. Take the task on 1080p images as an example, the PSNR of the LPTN is just reduced from $22.09$ to $21.95$ when the $L$ is increased from $3$ to $5$, yet the model achieves a speed-up of more than $\times 2$ and takes about $1/16$ of memory usage. This result validates that domain-specific attributes are presented in a relatively low-dimensional space.
	
	\subsection{Visual Comparisons}
	\label{visual_comp}
	
\begin{figure*}[t]
		\centering
		\includegraphics[width=0.98\textwidth]{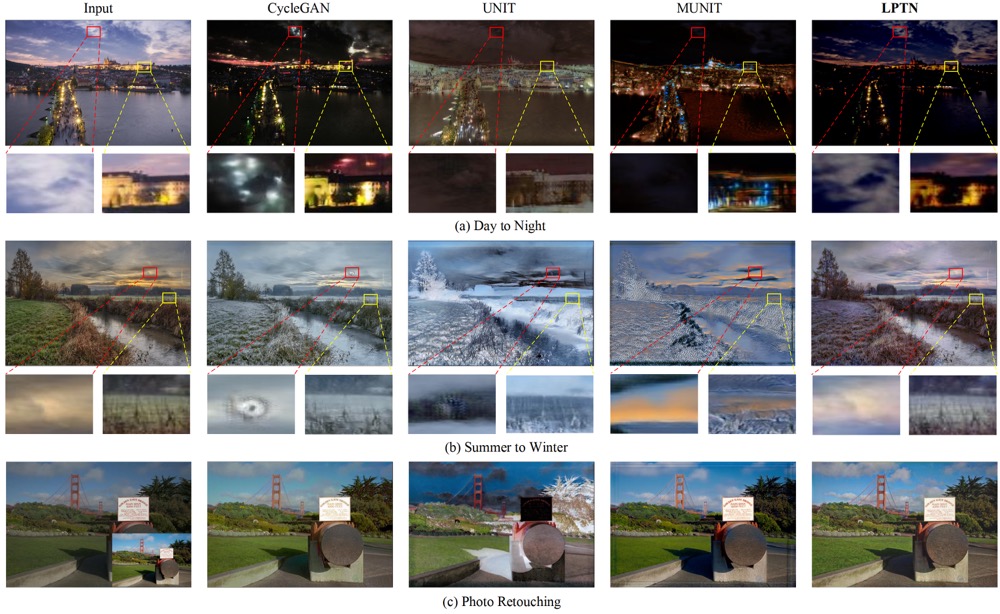}
		\caption{Visual comparisons among different I2IT methods, \ie, CycleGAN, UNIT, MUNIT and the proposed LPTN, on three different I2I tasks. The red and yellow boxes in (a) and (b) zoom in the particular regions for a better observation.
			\label{comparison_I2I}}
	\end{figure*}

	\noindent\textbf{Photorealistic I2IT: } We compare the visual performance on various photorealistic I2IT tasks, \ie, (a) day$\rightarrow$night, (b) summer$\rightarrow$winter and (c) photo retouching, in Figure~\ref{comparison_I2I}. This experiment is conducted on 1080p resolution considering the memory limitation of the CycleGAN, UNIT, and MUNIT. As shown in the figure, the proposed LPTN performs favorably against these three methods on both the photorealism and translation performance, while the LPTN is the only one that can be extended to higher resolution tasks (\eg, 4K).
	
	In specific, for the day$\rightarrow$night task as shown in Figure~\ref{comparison_I2I} (a), the LPTN translates the inputted day image into a dark night and shows little texture distortion. The geometric structure of the zoomed-in regions, \ie, a part of clouds and building, is well preserved in the translated results. Meanwhile, the global tone of the image is modified into a dark night style. The CycleGAN, which also achieves a dark tone, shows the second-best performance among these methods. However, it introduces many visible distortions, \eg, the cloud in the red box is transformed into many light spots while the ambient sky is in pure black. There are also some artifacts on top of the building as shown in the yellow box. The structural distortions and artifacts in the results of CycleGAN may be caused by the insufficient reconstruction capability of the decoder given a relatively high-resolution application. In contrast, LPTN achieves the encoding-decoding process via a closed-form filtering, which can be extended to higher resolutions, \eg, 4K, with negligible performance reduction. Similar conclusions can be made on the (b) summer$\rightarrow$winter and the (c) photo retouching tasks.
	
	We compare the proposed LPTN with traditional I2IT methods, \ie, CycleGAN, UNIT, and MUNIT, to demonstrate the advantages of our method. Generally, traditional ones are based on auto-encoder frameworks with mainly three steps: 1) disentangling the contents and attributes on a low-dimensional latent space via an encoding process; 2) translating the latent attribute code via residual blocks; 3) reconstructing the image from the translated attribute code via a decoder mirroring the encoding process. Actually, the ability to reconstruct contents is modeled by the network parameters of the auto-encoder. As a result, these methods can hardly be extended to high-resolution tasks or be applied to photorealistic scenarios due to the expensive computational cost.
	
	Instead of a parameterized encoding and decoding framework, the proposed LPTN decomposes the image into different frequency bands with tapering resolutions via a closed-form operation. The decomposed components are validated to be effective to represent the domain-specific attributes and content textures (as shown in Figure~\ref{motivation}). Consequently, the image can be easily reconstructed in a closed-form (note that the decomposition and reconstruction cost less than $2ms$ for a 4K image with $L=4$). As shown in Figure~\ref{pipeline}, most computation resources are allocated to translate the low-frequency component at the smallest resolution and to calculate the adaptive mask at the second-smallest resolution. Therefore, the proposed LPTN can be easily extended to higher resolution applications with linear growth of time consumption.
	
	Considering the inherent property of the Laplacian pyramid, the proposed LPTN cannot handle the problem generating novel content details, \eg, synthesizing Cityscapes images from its semantic segmentation labels. Actually, existing methods such as pix2pix perform well on this task by modeling the visual contents in a deep network, which depends on pixel-wise supervision and have a drastic demand for computational resources. A major limitation of our method is about the processing of high frequency (HF) components. Our progressive masking strategy saves much computation but may introduce halo artifacts in the day to night task. A feasible solution is to leverage the sparsity property of HF components, and employ sparse convolution on HF components to achieve more flexible translation while maintaining high efficiency.
	
	\subsection{Quantitative Examinations}
	\label{quantitative_comp}
	
	\begin{table}[t]
		\caption{Quantitative comparison on the MIT Adobe FiveK dataset (the photo retouching task). The N.A. denotes that the result is not applicable due to the limitation of computational resources.}
		\small
		\label{psnr_enhance}
		\begin{center}
			\begin{tabular}{p{2.3cm}p{0.5cm}<{\centering}p{0.5cm}<{\centering}p{0.5cm}<{\centering}p{0.5cm}<{\centering}p{0.5cm}<{\centering}p{0.5cm}<{\centering}}
				\toprule
				\multirow{2}*{Methods}&\multicolumn{2}{c}{480p}&\multicolumn{2}{c}{1080p}&\multicolumn{2}{c}{original}\\
				& PSNR &  SSIM & PSNR & SSIM& PSNR & SSIM \\
				\midrule
				CycleGAN~\cite{zhu2017unpaired} & 20.98 & 0.831  & 20.86 & 0.846& N.A. & N.A. \\
				UNIT~\cite{liu2017unsupervised} & 19.63 & 0.811  & 19.32 & 0.802 & N.A. & N.A. \\
				MUNIT~\cite{huang2018multimodal} & 20.32 & 0.829  & 20.28 & 0.815& N.A. & N.A. \\
				White-Box~\cite{hu2018exposure} & 21.32 &  0.864 & 21.26 & 0.872& 21.17 & 0.875 \\
				DPE \cite{chen2018deep}& 21.99 &  0.875 & 21.94 &0.885 & N.A. & N.A. \\
				\midrule
				\textbf{LPTN}, $ L=3 $ & 22.12 & 0.878  & 22.09 & 0.883 & 22.02 & 0.879 \\
				\textbf{LPTN}, $ L=4 $ & 22.10 & 0.872  & 22.03 & 0.870& 21.98 & 0.862 \\
				\textbf{LPTN}, $ L=5 $ & 21.94 & 0.866  & 21.95 & 0.858& 21.89 & 0.862 \\
				\bottomrule
			\end{tabular}
		\end{center}
	\end{table}	
	
	\begin{table}[t]
		\caption{Comparison about the time consumption (in seconds) of different inference models. Each result is an average of $50$ tests, where the N.A. denotes that the method cannot handle the image of specific resolution on a GPU with $11$G RAM.
		}
		\vspace{-1.2em}
		\small
		\label{time_comparison}
		\begin{center}
			\begin{tabular}{p{2.3cm}p{1cm}<{\centering}p{1cm}<{\centering}p{1cm}<{\centering}p{1cm}<{\centering}}
				\toprule
				Methods&480p&1080p&2K&4K\\
				\midrule
				CycleGAN~\cite{zhu2017unpaired}  &0.325&0.562&N.A.&N.A. \\
				UNIT~\cite{liu2017unsupervised} &0.294&0.483&N.A.&N.A. \\
				MUNIT~\cite{huang2018multimodal} &0.336&0.675&N.A.&N.A. \\
				White-Box~\cite{hu2018exposure} &2.846&5.123&6.542&9.785 \\
				DPE \cite{chen2018deep}&0.032&0.091&N.A.&N.A. \\
				\midrule
				\textbf{LPTN}, $ L=3 $ &0.003&0.012&0.043&0.082 \\
				\textbf{LPTN}, $ L=4 $ &0.002&0.007&0.015&0.033 \\
				\textbf{LPTN}, $ L=5 $ &0.0008&0.005&0.011&0.016 \\
				\bottomrule
			\end{tabular}
		\end{center}
	\end{table}

	In this section, we quantitatively compare the LPTN to the state-of-the-art methods on photo retouching regarding the PSNR/SSIM and time consumption.

	\noindent\textbf{Performance: }
	To test the performance on matching the manually retouched targets, we conduct three groups of experiments with the resolution being 480p, 1080p and original size (ranging from $ 3000\times 2000 $ to $ 6000\times 4000 $), respectively. As shown in Table~\ref{psnr_enhance}, the proposed LPTN performs favorably against both the general I2IT and the photo retouching methods. For the photo retouching task defined in the fiveK dataset, the main difference between the inputs and targets lies in the global tone (regarding colors or illuminations, \etc) of the image. The general I2IT methods translate the global tone satisfactorily yet perform badly on reconstructing the details as shown in Figure~\ref{comparison_I2I} (c). The main reason is that the fiveK dataset is relatively small but contains various scenes in the testing set so that the decoder can hardly learn a reverse mapping against the encoder on all visual scenes. For the photo retouching methods such as DPE~\cite{chen2018deep}, in contrast, a skip connection between the input and output is added to improve the reconstruction performance. However, the connection may also bring the unaesthetic visual attributes of the input images to the outputs caused by an unsatisfactory disentanglement of domain-invariant contents and domain-specific attributes. Thanks to the full decomposition and the preservation of reconstruction capacity by adaptively refining the high-frequency components, the proposed LPTN performs well on the photo retouching task.
	
	\noindent\textbf{Running Time: } As shown in Table~\ref{time_comparison}, the proposed LPTN outperforms other methods regarding the time consumption performance by a large gap, \eg, achieves about $\times 80$ speed-up against the CycleGAN on 1080p images when $ L=4 $, and runs on 4K images in real-time when $L=5$. According to Figure~\ref{pipeline}, the main optimization-based computations of the proposed method are concentrated on translating the low-frequency component $I_L$ and learning the mask for the last high-frequency component $h_{L-1}$, where both $I_L$ and $h_{L-1}$ are of low-resolution. For example, to translate an 1080p image ($I_0\in\mathbb{R}^{1920\times 1080\times 3}$) with $L=4$, we have $I_L\in\mathbb{R}^{120\times 67\times 3}$ and $h_{L-1}\in\mathbb{R}^{240\times 135\times 3}$. Besides, thanks to the spatial correlations among the high-frequency components, the generation of higher-resolution masks is efficient since they only include a bilinear interpolation operation and two convolutional layers.

	\subsection{User Study}
	
	To evaluate the overall performance of the translation regarding both the photorealism and transformation effects, we perform a user study based on human perception. In specific, we randomly select $20$ samples for the photorealistic day$\rightarrow$night and summer$ \rightarrow $winter translation tasks, respectively, and collect the translated results of the compared methods. A group of $ 20 $ participants are required to answer the following two questions after seeing the inputted images and all the compared results: 1) Photorealism: given the input image, which result is the most realistic one? 2) Transformation effectiveness: given the input image, which result is translated to the target style mostly? The results are summarized in Table~\ref{user_study_I2I}. For example, the proposed LPTN achieves a score of $78.3\%$ and $50.2\%$ for the visual performance of photorealism and transformation effect on the day$\rightarrow$night translation task, respectively. The results demonstrate that the proposed method performs better in preserving the content details and translating the images into target styles. The other three methods do not perform well on this subjective task since there are visible structural distortions and artifacts of their results. Some participants ($22.5\%$) prefer the output of CycleGAN regarding the transformation effect. Such preference mainly happens in those scenes that do not contain abundant detail textures, \eg, scenes consisting of a large area of sky or sea. Similar performance can be found in the summer$ \rightarrow $winter translation task.

	\begin{table}[t]
		\caption{User preference toward photorealistic day$ \rightarrow $night translation task. Participants are required to select out the most realistic and aesthetically pleasing result among the four methods. The images are shown in random order in each test.
		}
		\small
		\vspace{-0.5em}
		\label{user_study_I2I}
		\begin{center}
			\begin{tabular}{p{1.9cm}p{1.3cm}<{\centering}p{0.9cm}<{\centering}p{1cm}<{\centering}p{0.8cm}<{\centering}}
				\toprule
				Visual Metrics&CycleGAN & UNIT & MUNIT & \textbf{LPTN}\\
				\midrule
				Photorealism&16.4\%&2.3\%&3.0\%&78.3\%\\
				Aesthetic&21.3\%&12.7\%&8.5\%&57.5\%\\
				\bottomrule
			\end{tabular}
		\end{center}
	\end{table}
	
	\section{Conclusion}
	
	We proposed an highly-efficient framework for the photorealistic I2IT problem, which significantly reduces the computational burden when handling high-resolution images while simultaneously keeping the transformation performance. By using the Laplacian pyramid to decompose the input image, we disentangled the domain-specific visual attributes and the textures with tapering resolutions in an invertible manner and learned the translation and refinement networks on low-resolution components. A progressive masking strategy was then developed to adaptively refine the high-frequency components in order to generate a photorealistic result. The so-called Laplacian pyramid translation network (LPTN) was applied to a set of photorealistic I2IT tasks, exhibiting not only a much faster running speed but also comparable or superior translation performance. In particular, LPTN can run at real-time on 4K resolution images by using a desktop GPU.

{\small
\bibliographystyle{ieee_fullname}
\bibliography{LPTNbib.bib}
}

\end{document}